\documentclass{article}

\PassOptionsToPackage{numbers}{natbib}
\usepackage[preprint]{neurips_2026}
\makeatletter
\renewcommand{\@noticestring}{}
\makeatother

\usepackage[utf8]{inputenc}
\usepackage[T1]{fontenc}
\usepackage{hyperref}
\usepackage{booktabs}
\usepackage{graphicx}
\usepackage{xcolor}
\usepackage{kotex}
\usepackage{subcaption}
\usepackage{amsmath}
\usepackage{amssymb}

\title{{\fontsize{15}{19}\selectfont AXE: AlpaSim E2E Closed Loop Challenge Technical Report}}

\newcommand{\asep}{\hspace{0.9em}}

\author{%
  \begin{minipage}{0.97\linewidth}
  \centering\normalfont\small
  Joonhee Lim$^{\dagger}$\asep Seonmyeong Lee$^{\ddagger}$\asep Sihwan Hwang$^{\ddagger}$\asep Hanbin Jang$^{\ddagger}$\asep Hiwon Shin$^{\ddagger}$\asep Dogun Kim$^{\ddagger}$\\[2pt]
  Hyelin Jo$^{\ddagger}$\asep Hyeonjun Jeong$^{\ddagger}$\asep Yongjae Lee\asep Junseong Kwak\asep Junhyeok Park\asep Satyam Mishra\\[2pt]
  Juyeb Shin\asep Uisung Kim\asep Hyeonseo Kim\asep Saparkhan Kassymbekov\asep Hailemichael Molla Gashaw\\[2pt]
  Abi Rahman Syamil\asep Mingyu Kim\asep Hyeongjun Kim\asep Yeina Lee\asep Moogeun Park\asep Dongsuk Kum$^{*}$\\[10pt]
  {\normalsize KAIST AXE Lab}\\[4pt]
  {\small \texttt{kingear3@kaist.ac.kr}$^{\dagger}$\quad \texttt{dskum@kaist.ac.kr}$^{*}$}\\[2pt]
  {\footnotesize $^{\dagger}$Project lead\quad $^{\ddagger}$Part lead\quad $^{*}$Corresponding author}
  \end{minipage}%
}

\begin{document}
\maketitle

\begin{abstract}

This report presents a comprehensive pipeline for developing an end-to-end (E2E) autonomous driving model for the Physical AI AV Track. The framework consists of data curation, HD map auto-labelling, data-driven imitation learning, and simulator-based reinforcement learning. To ensure safety in the closed-loop setting, we introduce a selection-based E2E planning approach that leverages object detection and BEV segmentation to filter out trajectory candidates likely to cause collisions or off-road driving. Trained on NVIDIA NuRec and the NVIDIA PAI-AV dataset, our method scores a fixed vocabulary of trajectory candidates for collision, drivable-area compliance, and progress learned from simulated rollouts, and selects the best candidate that passes the filter. This design highlights the effectiveness of combining E2E planning with perception-based trajectory filtering for safe and reliable closed-loop autonomous driving.

\end{abstract}

\section{Method summary}

\begin{figure}[t]
    \centering
    \includegraphics[width=1.00\linewidth]{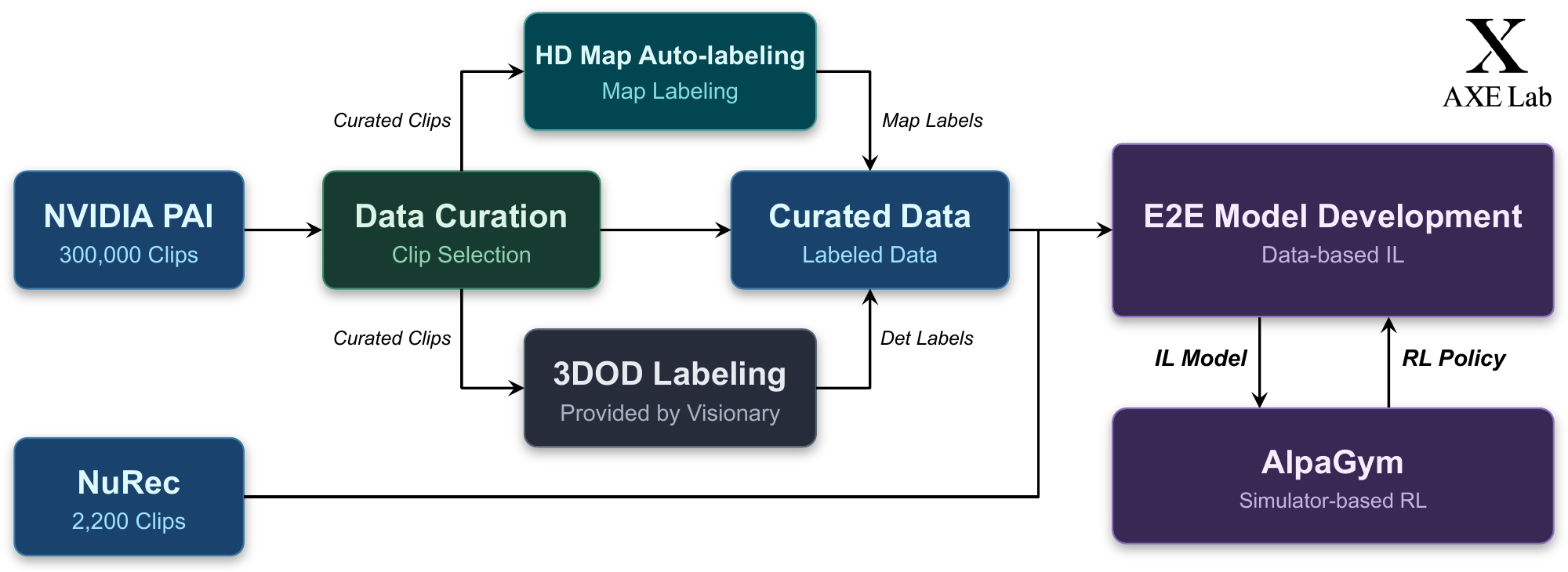}
    \caption{Comprehensive workflow for end-to-end model development.}
    \label{fig:system_architecture}
\end{figure}

Our proposed method for the Physical AI AV Track comprises an end-to-end (E2E) autonomous driving pipeline designed to achieve robust performance in closed-loop execution. As illustrated in Figure~\ref{fig:system_architecture}, the framework follows a sequential development and training pipeline comprising five core stages, from data curation and imitation learning to reinforcement learning-based policy refinement.

\begin{enumerate}
    \item \textbf{Data Curation:} We curate 1,400 additional PAI-AV clips using scenario-based, semantic, and spatiotemporal feature-based selection to address coverage gaps and model failures.  The selected clips are subsequently auto-labeled and used for end-to-end policy training.

    \item \textbf{HD Map Auto-labeling:}
    HD map annotations for the PAI-AV dataset are automatically generated using a 3DGS-based method.
    We extend PseudoMapTrainer~\cite{lowens2025pseudomaptrainer} with LiDAR supervision and DINOv3~\cite{simeoni2025dinov3} distillation to ensure annotation quality.

    \item \textbf{End-to-End Autonomous Driving Model:} We develop an E2E architecture that encodes multi-view temporal observations into a BEV representation and selects a safe trajectory from 8,192 candidates via perception-based feasibility checks and coarse-to-fine scoring.

    \item \textbf{Data-driven Imitation Learning:} We train the E2E model on NVIDIA NuRec data and auto-labeled PAI-AV data using perception, imitation, and scoring losses.

    \item \textbf{Simulator-based Reinforcement Learning:} Finally, to address compounding errors and enhance robustness in out-of-distribution states, we fine-tune the policy using reinforcement learning within a realistic closed-loop simulator, NVIDIA AlpaSim.

\end{enumerate}

\section{Policy description}

\subsection{Data Curation}
    We curate complementary driving clips to address scenario imbalance and model failures while limiting map-refinement effort. Starting from approximately 2K NuRec clips, we select additional data from a labeled 10K-clip subset of NVIDIA Physical AI (PAI), which contains approximately 300K clips overall. We retain the relatively accurate PAI object annotations and refine map annotations for selected clips before training. 
To select informative clips for downstream policy training, we apply three complementary selection strategies and select a total of 1{,}400 PAI-AV clips.

\paragraph{Scenario-based Selection.}
We characterize clips using longitudinal and lateral ego motion and the distribution of surrounding objects to identify scenarios underrepresented in NuRec. We initially select clips to improve scenario coverage, including rare lane-change maneuvers. After initial model training, we select clips resembling low-scoring or failed scenarios, including turning maneuvers and high-speed straight driving.

\paragraph{VLM-guided Scene Selection.}
To capture context beyond predefined scenario labels, we use a pretrained VLM to generate scene descriptions and tags. These guide the selection of clips with contexts similar to model failures, while visual inspection helps identify rare cases for fine-tuning.

\paragraph{Scene-embedding-based Selection.}
We further select clips using distances between spatiotemporal scene embeddings extracted by SparseDrive. A hierarchical scene encoder aggregates sparse object and map tokens within each frame and models their temporal evolution to obtain clip-level embeddings. These embeddings guide selection based on novelty relative to existing training data and similarity to failure scenarios.

\subsection{HD Map Auto-labeling}
    \label{sec:hd_map_auto_labeling}
We extend PseudoMapTrainer~\cite{lowens2025pseudomaptrainer} and have developed our own auto-labeling framework to generate HD map annotations for downstream driving model training.

\paragraph{Semantic Road Surface Reconstruction.}
PseudoMapTrainer only optimizes road surface Gaussians with perspective view RGB and semantic label images.
This may result in inconsistent reconstruction due to inconsistencies across the multi-view images.
To address such limitations, we leverage auxiliary supervision from static point cloud reconstruction using LiDAR observations.
In addition, we introduce DINOv3~\cite{simeoni2025dinov3} latent embedding to each Gaussian representation to boost multi-view consistency.

\paragraph{Semantic HD Map Pseudo-Label Generation.}
The resulting semantically enhanced road Gaussian representation is rendered into ego-centric semantic Bird's-Eye-View (BEV) maps, which serve as pseudo-labels for downstream driving model training.

\subsection{End-to-End Autonomous Driving Model}
    \label{ss_2_3}
Our method formulates trajectory planning as selection from a fixed vocabulary of 8,192 candidate trajectories. The model encodes multi-view temporal observations into a bird's-eye-view (BEV) representation and leverages perception outputs to filter infeasible candidates based on drivable-area constraints and object collisions, promoting safe driving. The remaining candidates are ranked through a coarse-to-fine scoring pipeline~\cite{yao2026drivesuprim} to select a single trajectory for closed-loop execution.

\paragraph{Inputs and BEV Encoding.}
The model takes three temporal frames ($t-1.0$, $t-0.5$, and $t$) from three front and front-side cameras ($512 \times 256$), the current frame from the two rear cameras ($512 \times 256$ each), along with ego state and 20 route waypoints. A DINOv3-pretrained ViT-S backbone~\cite{simeoni2025dinov3} and a three-layer BEVFormer encoder~\cite{li2024bevformer} produce a $112 \times 112$ BEV grid, where ego-motion-aligned temporal features are fused via convolution. The two rear views are encoded separately by the same ViT-S, without temporal fusion or BEV projection, and pooled into 256 rear tokens.

\paragraph{Trajectory Scoring and Selection.}
A three-layer transformer decoder scores 8,192 candidate trajectories from a fixed $k$-means trajectory vocabulary using $56 \times 56$ pooled BEV tokens, the 256 rear tokens, route features, and ego state. The scoring heads predict no-fault collision (NC), drivable-area compliance (DAC), ego progress (EP), their product (the overall scene score), and imitation similarity. A coarse-to-fine pipeline~\cite{yao2026drivesuprim} ranks all candidates and re-scores the top 256 with a separate refinement decoder over the same pooled BEV features and rear tokens. A perception-based feasibility gate, derived from perception heads operating on the same BEV representation, penalizes candidates that violate drivable-area constraints or overlap with detected objects. The highest-ranked candidate is output as a sequence of 40 poses $(x,y,\theta)$ at 0.1\,s intervals over a 4\,s horizon and tracked by an MPC controller.

\subsection{Data-driven Imitation Learning}
    During the imitation learning stage, the E2E model is trained on real-world driving logs using joint perception and planning objectives to learn scene understanding and trajectory selection. Focal loss~\cite{lin2017focal} is used for 3D object classification, L1 loss with Hungarian matching~\cite{carion2020end} for bounding-box regression, and class-weighted cross-entropy for BEV segmentation. Soft cross-entropy is used to imitate recorded trajectories, while auxiliary losses predict simulator-generated NC, DAC, and EP labels and their product. Both planning losses are applied to all 8,192 candidates at the coarse stage and the top 256 at the fine stage. Further details are provided in Appendix~\ref{app_il}.

\subsection{Simulator-based Reinforcement Learning}
    For online reinforcement learning (RL), we utilize the 3D Gaussian Splatting (3DGS)-based simulator, $\text{NVIDIA AlpaSim}$. To mitigate the computational overhead of 3DGS rendering and alleviate long-horizon credit assignment challenges, the RL agent interacts with the environment at a downsampled frequency of $2\text{ Hz}$, derived from the native $10\text{ Hz}$ simulation rate. We employ Proximal Policy Optimization (PPO)~\cite{schulman2017proximal} for online policy optimization. The episode reward decomposes into a step-wise function $r_t = \mathrm{RC}_t \cdot \mathrm{LD}_t - \mathrm{Cor}_t - T_t$, where $\mathrm{RC}_t \in [0, 1]$ and $\mathrm{LD}_t \in [0, 1]$ denote route completion and normalized lateral alignment, respectively. $\mathrm{Cor}_t \in [0, 1]$ is the corridor penalty for lateral distances between $1\,\text{m}$ and $4\,\text{m}$. $T_t \in \{0, 1\}$ is a hard constraint penalty triggered by safety violations (e.g., collisions, off-road driving, or boundary cross-overs), which terminates the episode immediately upon violation.

\section{Resources and training}

\subsection{Data Curation}
    For additional training data, we use a labeled subset of approximately 10K clips from the NVIDIA PAI-AV dataset, which contains approximately 300K clips in total. 
We apply three complementary selection paradigms across four curation rounds: scenario-based coverage and failure-aware selection, semantic retrieval using Qwen-based scene metadata~\cite{qwen36_27b}, and feature-space selection using temporal scene representations extracted by SparseDrive~\cite{sun2025sparsedrive}.
The SparseDrive model used for feature extraction was trained on an approximately 1K-clip subset of the labeled PAI data. 
Following deduplication across the curation rounds, 1,400 unique PAI-AV clips were retained. Their map annotations were generated using the HD Map Auto-labeling pipeline described in Sec.~\ref{sec:hd_map_auto_labeling} and ~\ref{ss_3_2}, while the available object annotations were retained. The resulting labeled PAI-AV clips were combined with the NuRec clips for downstream
end-to-end policy training.
The selected datasets and models were used according to their respective usage terms and are listed in the references.

\subsection{HD Map Auto-labeling}
    \label{ss_3_2}

We generate HD map annotations on the curated NVIDIA PAI-AV clips using multi-view camera images, LiDAR point clouds, and ego poses. Our pipeline builds upon PseudoMapTrainer~\cite{lowens2025pseudomaptrainer} for road surface reconstruction using Gaussian representation. For semantic label for Gaussian optimization, we employ off-the-shelf Mask2Former~\cite{cheng2022masked}, trained on MapillaryVistasV2~\cite{neuhold2017mapillary} dataset, to extract semantic label predictions. We additionally use pretrained DINOv3~\cite{simeoni2025dinov3} ViT-L model to provide rich feature optimization.

\subsection{End-to-End Autonomous Driving Model}
    \label{ss_3_3}

The image backbone uses publicly available pre-trained DINOv3 weights~\cite{simeoni2025dinov3}, the BEV encoder follows the BEVFormer architecture~\cite{li2024bevformer}, and trajectory scoring is based on DriveSuprim~\cite{yao2026drivesuprim}.

\subsection{Data-driven Imitation Learning}
    \label{ss_3_5}

We train on NVIDIA NuRec 26.01 and 26.04 together with clips from the NVIDIA PAI-AV dataset. For NuRec, 3D bounding boxes and BEV drivable-area labels come from the provided annotations and maps. For PAI-AV, 3D bounding boxes are obtained by auto-labeling, and we additionally auto-label a BEV drivable-area raster for every frame, which serves both as the segmentation target and as the drivable area for DAC scoring. Planner labels are generated by tracking each of the 8,192 vocabulary trajectories with an MPC controller and scoring the rollout with NC, DAC, EP, and their product, with EP set to zero for lateral deviations exceeding 4\,m from the GT trajectory. Training proceeds in three stages: 20 epochs of perception-only training, 5 epochs with frozen backbone and BEV encoder, and 5 epochs of joint training. We use AdamW~\cite{loshchilov2017decoupled} with a learning rate of $10^{-4}$ and weight decay of 0.01; each stage starts with a 500-iteration linear warmup, followed by a cosine decay to $10^{-6}$ scheduled over 30 epochs and truncated at the end of the stage. Training uses a batch size of 64 across four NVIDIA H200 GPUs. Further details are provided in Appendix~\ref{app_il}.

\subsection{Simulator-based Reinforcement Learning}
    Closed-loop RL fine-tuning is conducted on AlpaSim using 1,761 training and 441 validation NuRec scenes across 48 parallel simulator sessions at a 2 Hz control frequency. Keeping the pretrained IL policy (48.3M parameters) fully frozen, we train only 0.54M LoRA~\cite{hu2021lora} parameters placed on planner's  transformer decoder layers and a separate 1.78M critic network. We fine-tune using PPO~\cite{schulman2017proximal} with an exact KL penalty ($\beta = 0.02$) relative to the SFT policy, optimized via AdamW~\cite{loshchilov2017decoupled} at a constant learning rate of $1 \times 10^{-4}$ and a rollout buffer batch size of 2,048 steps (minibatch size 512) for 17 epochs across $8\times$ NVIDIA H100 GPUs.

\section{Limitations and safety considerations}

\paragraph{Failure Modes.}
The planner selects from a fixed set of 8{,}192 trajectories clustered from
recorded driving, so maneuvers outside this set (e.g., tight U-turns or
multi-point turns) cannot be produced, and rare situations are covered by fewer
nearby candidates.

\paragraph{Limitations.}
Trajectories are scored and selected using collision avoidance,
drivable-area compliance, ego progress, and imitation. Ride comfort
(e.g., acceleration, jerk, and yaw rate) is not part of the selection criteria,
so the chosen trajectories can be less smooth than human driving.

\paragraph{Safety Considerations.}
The candidate filter relies on the model's own front-view BEV perception, so missed detections and rear agents, seen by the planner only through rear-camera tokens, can pass it.
Traffic-light state is not part of the planner labels, so the planner can proceed through a red light.

\section*{Acknowledgments}
We thank Visionary for providing 3D object annotations.
This research was supported by the Advanced GPU Utilization Support Program funded by the Government of the Republic of Korea (Ministry of Science and ICT).

\bibliographystyle{IEEEtran}
\bibliography{Ref}

\newpage
\appendix
\section{Additional details}

\subsection{Data Curation}\label{A.1}
\begin{figure}[!bp]
    \centering
    \includegraphics[width=0.98\textwidth]{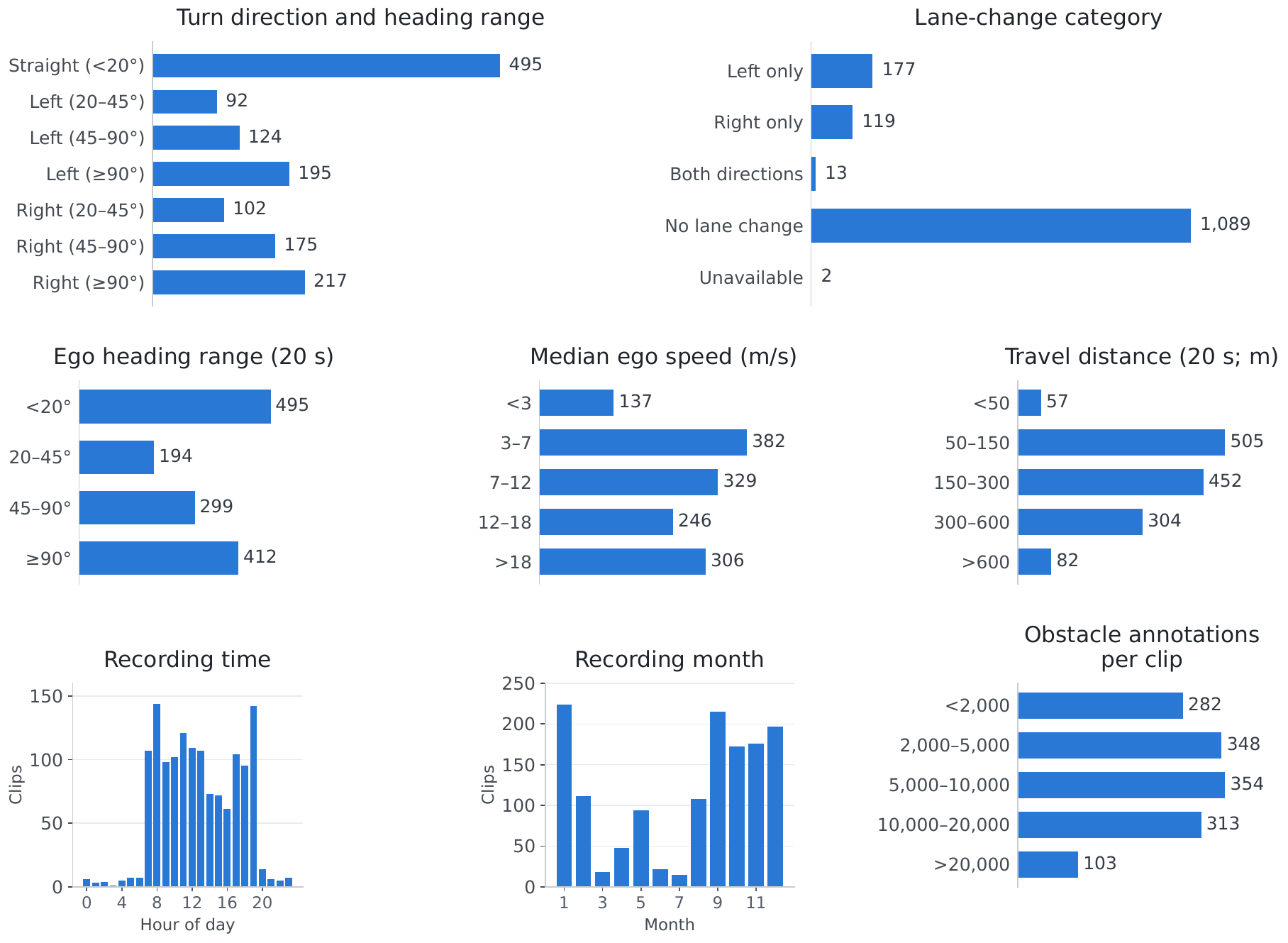}
    \caption{
        Scenario distribution of the PAI subset curated by data selection strategies. Each panel reports the 1,400 clips grouped by one clip-level attribute, and the numbers are clip counts.
    }
    \label{fig:scenario_plot}
\end{figure}

\begin{figure}[!bp]
    \centering
    \includegraphics[width=0.99\textwidth]{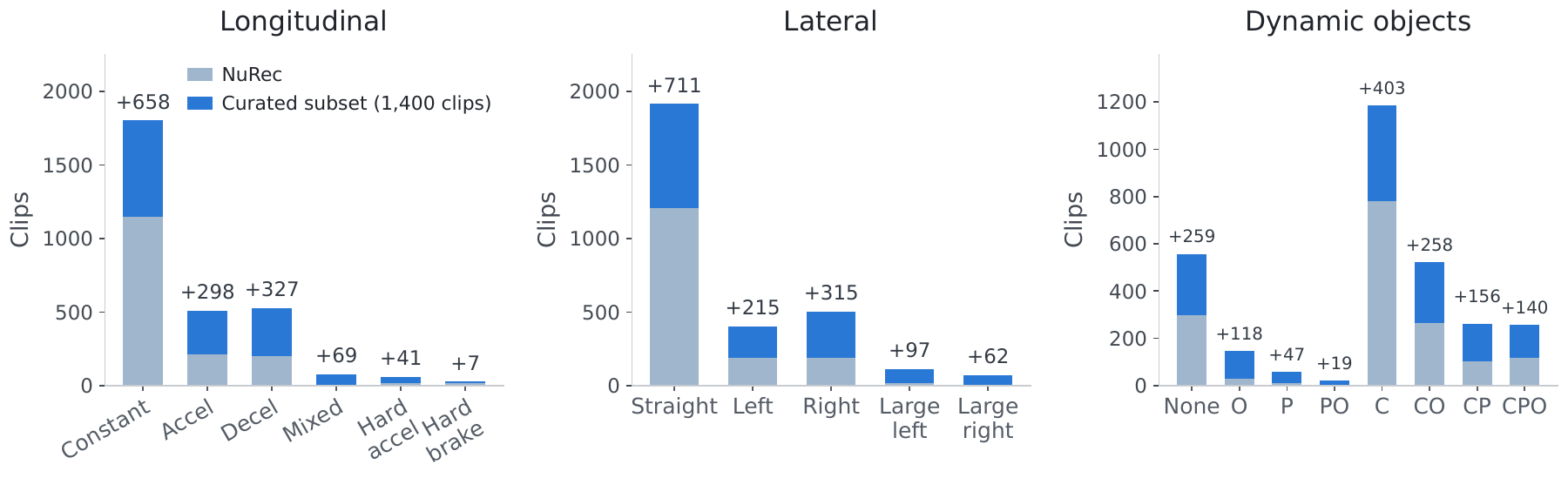}
    \caption{
        Scenario coverage of NuRec (Light blue) before and
        after compensation by the curated subset (Blue). Bar labels give the number of clips the
        curation adds to each class. The dynamic-object panel shows the eight
        combinations of vehicle (C), pedestrian (P) and other (O) presence.
    }
    \label{fig:nurec_vs_curated_plot}
\end{figure}

    We curate 1{,}400 PAI clips from a candidate pool of 9{,}924. The pool is built
in two steps: we first filter the full NVIDIA PAI release by country and
vehicle platform, so that calibration and traffic conventions stay consistent, and then apply data
selection. Every clip in the pool is then auto-labelled with our own pipeline for ego trajectory, objects and map information, so all curation operates on a common set of annotations. Three complementary strategies draw from this pool. Sec.~\ref{A.1.1} selects over explicit clip attributes, Sec.~\ref{A.1.2} retrieves over VLM semantic tags, and Sec.~\ref{A.1.3} selects in a learned latent space. Figure~\ref{fig:scenario_plot} summarizes the scenario distribution
of the curated subset, while Fig.~\ref{fig:nurec_vs_curated_plot}
shows how it supplements NuRec along the scenario axes defined in Sec.~\ref{A.1.1}.

\subsubsection{Scenario-based Data Selection}
\label{A.1.1}

Our first selection pass targets the gaps in NuRec rather than the PAI pool
itself. To make "gap" measurable we define a scenario taxonomy over three
independent axes, namely longitudinal ego behaviour, lateral ego behaviour and
the presence of surrounding dynamic objects, and label every clip on each axis
from the obstacle annotations and ego-motion information. The axes yield
$6\times5\times8 = 240$ scenario cells. Characterising NuRec in these terms
exposes which cells it covers thinly, and we then draw PAI clips from exactly
those cells.

\textbf{Longitudinal axis.}
Let $a_x(t)$ denote the longitudinal ego acceleration over a clip $[t_0,t_1]$,
and let $a^{+}_x(t)=\max\big(a_x(t),0\big)$ and $a^{-}_x(t)=\min\big(a_x(t),0\big)$
denote its positive and negative parts. We summarise the clip by two peaks and
two accumulated efforts,
\begin{equation}
\hat{a}^{+} = \max_{t\in[t_0,t_1]} a^{+}_x(t),
\quad
\hat{a}^{-} = \min_{t\in[t_0,t_1]} a^{-}_x(t),
\quad
I^{+} = \int_{t_0}^{t_1}\! a^{+}_x(t)\,dt,
\quad
I^{-} = -\int_{t_0}^{t_1}\! a^{-}_x(t)\,dt ,
\end{equation}
where $\hat{a}^{+}$ and $\hat{a}^{-}$ denote the peak acceleration and peak
deceleration, and $I^{+}$ and $I^{-}$ denote the total acceleration and
deceleration accumulated over the clip. Peaks capture brief, sharp events while
the integrals capture sustained effort, which separates a clip that accelerates
steadily from one whose speed merely drifts. Table~\ref{tab:long_scen} lists
the six behaviours each clip falls into under thresholds
$T_{\hat{a}^{+}}=3$ and $T_{\hat{a}^{-}}=-4$\,m/s$^2$ for the peaks and
$T_{I^{+}}=9$ and $T_{I^{-}}=8$\,m/s for the integrals, set by inspecting clips
at each value.

\textbf{Lateral axis.}
Yaw change per unit time confounds turning with speed, so we measure yaw change
over a fixed travelled \emph{distance} instead. Let $\psi(t)$ denote the ego
heading and $s(t)$ its position. For a window of $20$\,m,
\begin{equation}
\Delta\psi_{20\mathrm{m}} = \psi(t_2) - \psi(t_1)
\quad\text{s.t.}\quad
\sum\nolimits_{t_1}^{t_2-0.1}\big|s(t+0.1)-s(t)\big| = 20\,\mathrm{m},
\qquad
\Delta\psi = \max\big|\Delta\psi_{20\mathrm{m}}\big| ,
\end{equation}
where $\Delta\psi_{20\mathrm{m}}$ denotes the heading change accumulated over one
such window and $\Delta\psi$ denotes the largest of these over the clip. Table~\ref{tab:lat_scen} lists the five classes each clip is assigned by
$\Delta\psi$.

\textbf{Dynamic-object axis.}
Obstacle classes are grouped into vehicles (\texttt{automobile}), pedestrians
(\texttt{person}) and other objects (\texttt{heavy\_truck}, \texttt{trailer},
\texttt{bus}, \texttt{rider}, \texttt{stroller}, \texttt{protruding\_object},
\texttt{other\_vehicle}, \texttt{train\_or\_tram\_car}, \texttt{animal}). A group
counts as present if any of its instances falls inside a region of interest
around the ego at any point in the clip, taken as $4$\,m to either side, $30$\,m
ahead and $10$\,m behind. The three groups give $2^3=8$ combinations.

\begin{table}[tbp]
\centering
\begin{minipage}[t]{0.50\textwidth}
\centering
\caption{Longitudinal scenario classes.}
\label{tab:long_scen}
\vspace{1mm}
\begin{tabular}{cll}
\toprule
\# & Scenario & Criterion \\
\midrule
1 & Hard braking      & $\hat{a}^{-} < T_{\hat{a}^{-}}$ \\
2 & Hard acceleration & $\hat{a}^{+} > T_{\hat{a}^{+}}$ \\
3 & Accelerating      & $I^{+} \geq T_{I^{+}}$ \\
4 & Decelerating      & $I^{-} \geq T_{I^{-}}$ \\
5 & Mixed             & $I^{+} \geq T_{I^{+}},\; I^{-} \geq T_{I^{-}}$ \\
6 & Constant speed    & otherwise \\
\bottomrule
\end{tabular}
\end{minipage}\hfill
\begin{minipage}[t]{0.48\textwidth}
\centering
\caption{Lateral scenario classes.}
\label{tab:lat_scen}
\vspace{1mm}
\begin{tabular}{cll}
\toprule
\# & Scenario & Criterion \\
\midrule
1 & Straight    & $-30^\circ < \Delta\psi < 30^\circ$ \\
2 & Left turn   & $-100^\circ < \Delta\psi \leq -30^\circ$ \\
3 & Right turn  & $30^\circ \leq \Delta\psi < 100^\circ$ \\
4 & Large left  & $-180^\circ < \Delta\psi \leq -100^\circ$ \\
5 & Large right & $100^\circ \leq \Delta\psi < 180^\circ$ \\
\bottomrule
\end{tabular}
\end{minipage}
\end{table}

\subsubsection{VLM-assisted Data Selection}
\label{A.1.2}

The scenario taxonomy of Sec.~\ref{A.1.1} describes how the ego moves and what
surrounds it, but not what the scene \emph{is}: a construction detour, a school
bus unloading and an ambulance being followed all collapse into the same cell.
We therefore tag the remaining pool of $5{,}258$ clips with the pretrained
Qwen3.6-27B~\cite{qwen36_27b} model and retrieve against the tags.

\textbf{Fields and prompt.} The fields follow the scenario-categorisation tags
of ISO~34504, kept to what a single forward camera can judge over $20$\,s. With
binary fields for the interactions a driving policy would have to handle, a
free-text summary, a difficulty rating and a confidence, the schema has $27$
fields, returned as one JSON object per clip. The ten fields whose reading is
otherwise open are pinned by a plain-language definition in Table~\ref{tab:defs}.

\textbf{Inference.} The $32$ frames sampled across $20$\,s are passed
through the model's video pathway, where the vision encoder pairs adjacent
frames into $16$ temporal steps, so motion between them is legible.

\begin{figure}[!bp]
    \centering
    \includegraphics[width=0.95\textwidth]{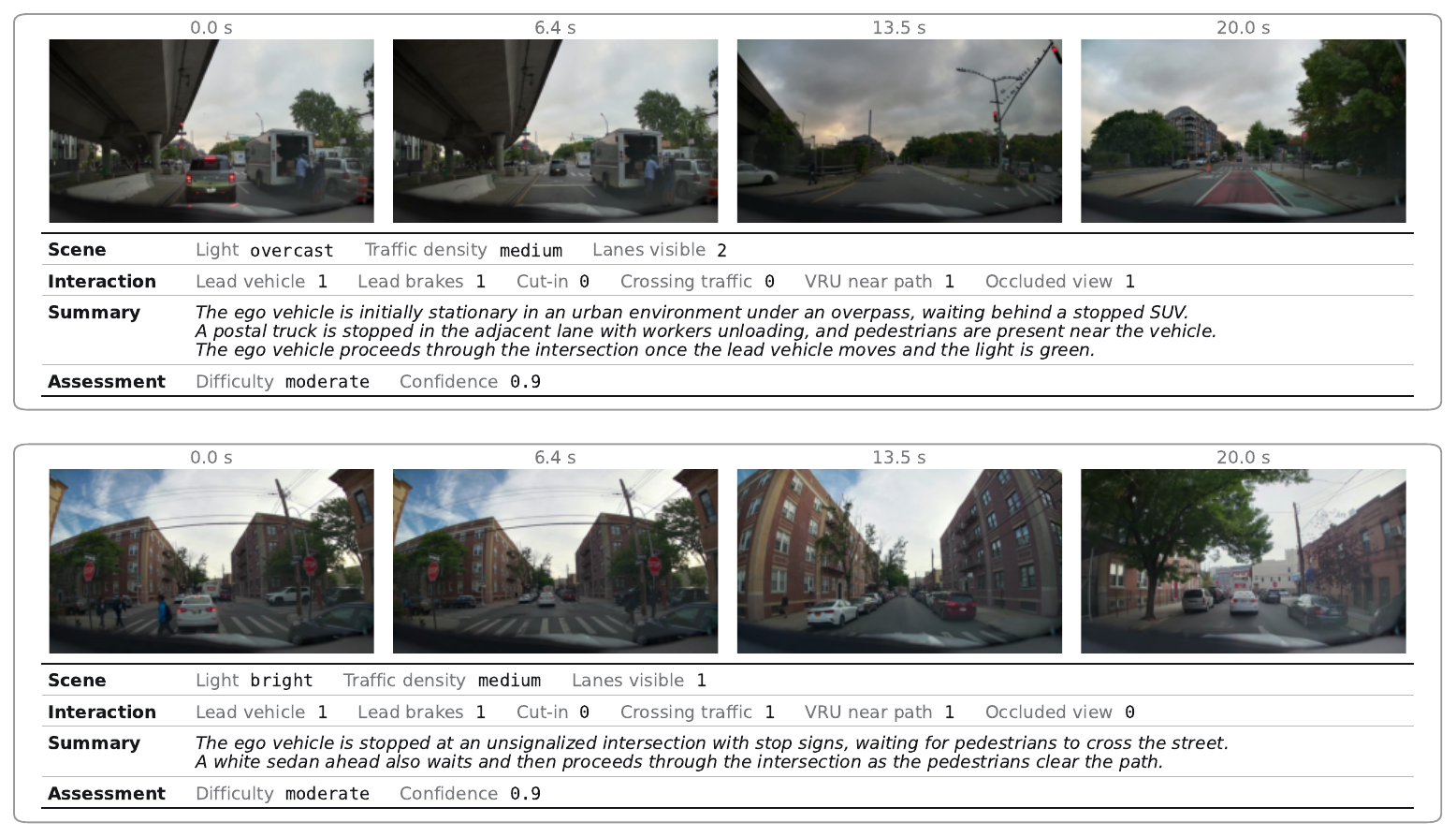}
    \caption{
        Two examples of the selected clips with our tag fields. Each panel shows the sampled frames given to the model, with the values it returned listed below them.
        (Top) A signalized crossing on an urban street under an elevated structure, with a postal truck stopped in the adjacent lane, and a lead vehicle braking ahead. (Bottom) An unsignalized crossing on an urban street, with pedestrians walking across the ego's path, parked cars along kerbs and a lead vehicle waiting at the stop line.
    }
    \label{fig:vlm_example}
\end{figure}

\textbf{Selection.} We query the tags in two ways. The first looks for clips
whose tagged context resembles the scenarios the model fails on. The second
looks for rare cases through tag combinations and the summaries. Candidates are
reviewed as contact sheets showing the sampled frames with their tags beneath,
and what we keep follows scarcity. Every accident or breakdown clip in the pool is taken and
nearly every emergency-vehicle one, while the abundant VRU and construction
categories are sampled sparsely. Figure~\ref{fig:vlm_example} shows two of the
clips kept this way. Although the taxonomy of Sec.~\ref{A.1.1} places them in the same cell, both accelerating and straight, the summaries using VLM separate them on semantics, which is what lets the selection retrieve by scene context rather than by attributes alone. One ego waits at a signal with a stopped truck blocking its view before pulling away, while the other pulls away only once pedestrians have cleared a stop-sign crossing.

\begin{table}[!tp]
\centering
\caption{The ten fields pinned by a definition in the prompt.
The rest are constrained by their permitted values alone. $^{*}$ denotes a
binary field, marked 1 when the event occurs at any point in the clip, which
makes it a clip-level judgement rather than a per-frame one.}
\label{tab:defs}
\vspace{1mm}
\begin{tabular}{lp{0.7\textwidth}}
\toprule
Field &  Definition as given to the model \\
\midrule
\texttt{light} & bright = clear daylight, overcast = flat grey, dim\_dusk = low sun or twilight with the sky still lit, dark\_shadow = tunnel or deep shade \\
\addlinespace[1pt]
\texttt{traffic\_density} & low $<3$ vehicles in view, medium $3$--$9$, high $\geq10$, over the clip \\
\addlinespace[1pt]
\texttt{lanes\_visible} & lanes in the ego's direction of travel \\
\addlinespace[0.8pt]
\texttt{lead\_vehicle}$^{*}$ & a vehicle directly ahead in the ego lane at any point \\
\addlinespace[0.8pt]
\texttt{occluded\_view}$^{*}$ & a vehicle or structure hides a meaningful part of the forward view \\
\addlinespace[0.8pt]
\texttt{cut\_in}$^{*}$ & another vehicle moves into the ego lane ahead \\
\addlinespace[0.8pt]
\texttt{lead\_brakes}$^{*}$ & the lead vehicle visibly brakes or slows sharply \\
\addlinespace[0.8pt]
\texttt{vru\_near\_path}$^{*}$ & a pedestrian or cyclist is in, or moving into, the ego's driving corridor \\
\addlinespace[0.8pt]
\texttt{crossing\_traffic}$^{*}$ & a vehicle crosses the ego's path, typically at a junction \\
\addlinespace[0.8pt]
\texttt{difficulty} & how demanding this clip would be for a driving policy \\
\bottomrule
\end{tabular}
\end{table}

\subsubsection{Spatiotemporal Feature Diversity}
\label{A.1.3}
Traditional rule-based scenario classification is limited by its inability to distinguish between discrete clips assigned to the same semantic class (e.g., straight driving, accelerating, or environments containing only vehicles). While suitable for initial data curation, this approach necessitates expansion into a continuous latent space for finer granularity.
To address this, we use a scene encoder to extract clip-level spatiotemporal embeddings, which enable the identification and selection of novel and diverse driving clips.

\begin{figure}[tbp]
    \centering
    \includegraphics[width=0.5\linewidth, trim={1mm 2mm 1mm 9mm}, clip]{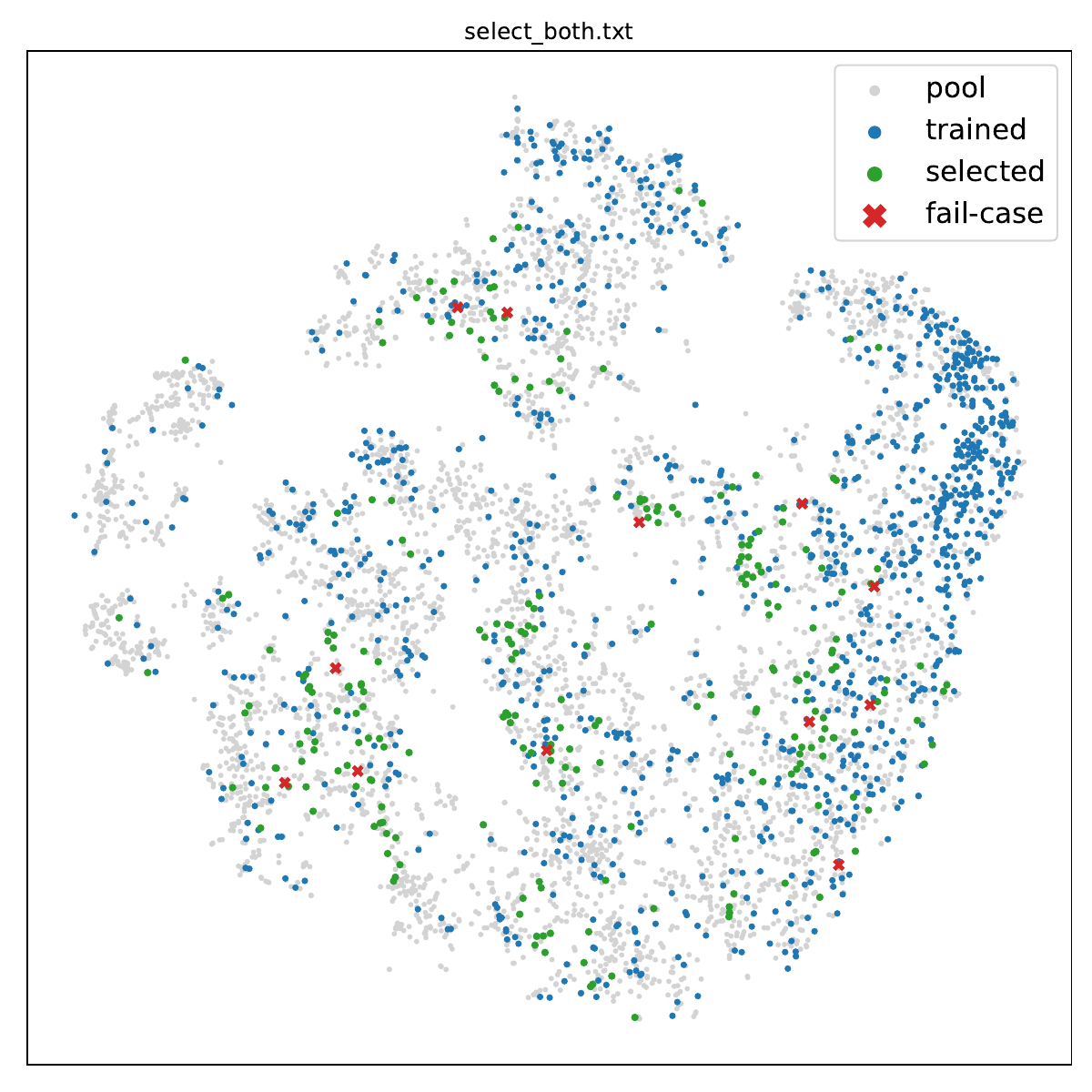}
    \caption{
        Illustrative visualization of an intermediate clip-selection run in the latent space. Colors indicate training clips (blue), failed evaluation clips (red), candidate clips (gray), and selected clips (green). The displayed selection is an intermediate diagnostic and does not represent the final number of curated clips.
    }
    \label{fig:latent_split}
\end{figure}

Figure~\ref{fig:latent_split} illustrates the sample selection process utilizing these latent embeddings. By leveraging the indices of previously trained and failed evaluation cases, we use the latent space distance to identify new samples that exhibit novel features and can potentially mitigate known failure modes. Specifically, we first filter for novelty by isolating clips that are distant from the existing training distribution, and subsequently prioritize clips situated in close proximity to the failed cases.

\begin{figure}[tbp]
    \centering
    \begin{subfigure}[b]{0.32\textwidth}
        \centering
        \includegraphics[width=\textwidth, trim={0mm 0mm 0mm 7mm}, clip]{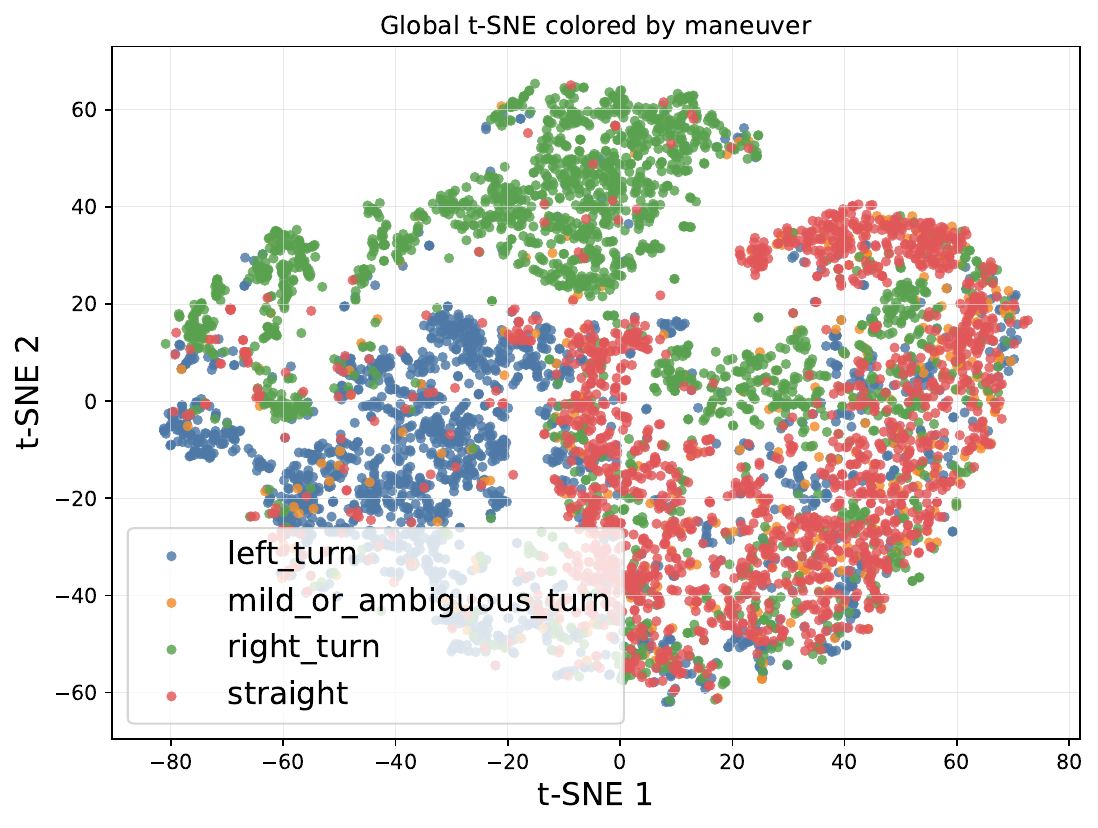}
        \caption{Lateral motion}
        \label{fig:latent_lateral}
    \end{subfigure}
    \hfill
    \begin{subfigure}[b]{0.32\textwidth}
        \centering
        \includegraphics[width=\textwidth, trim={0mm 0mm 0mm 7mm}, clip]{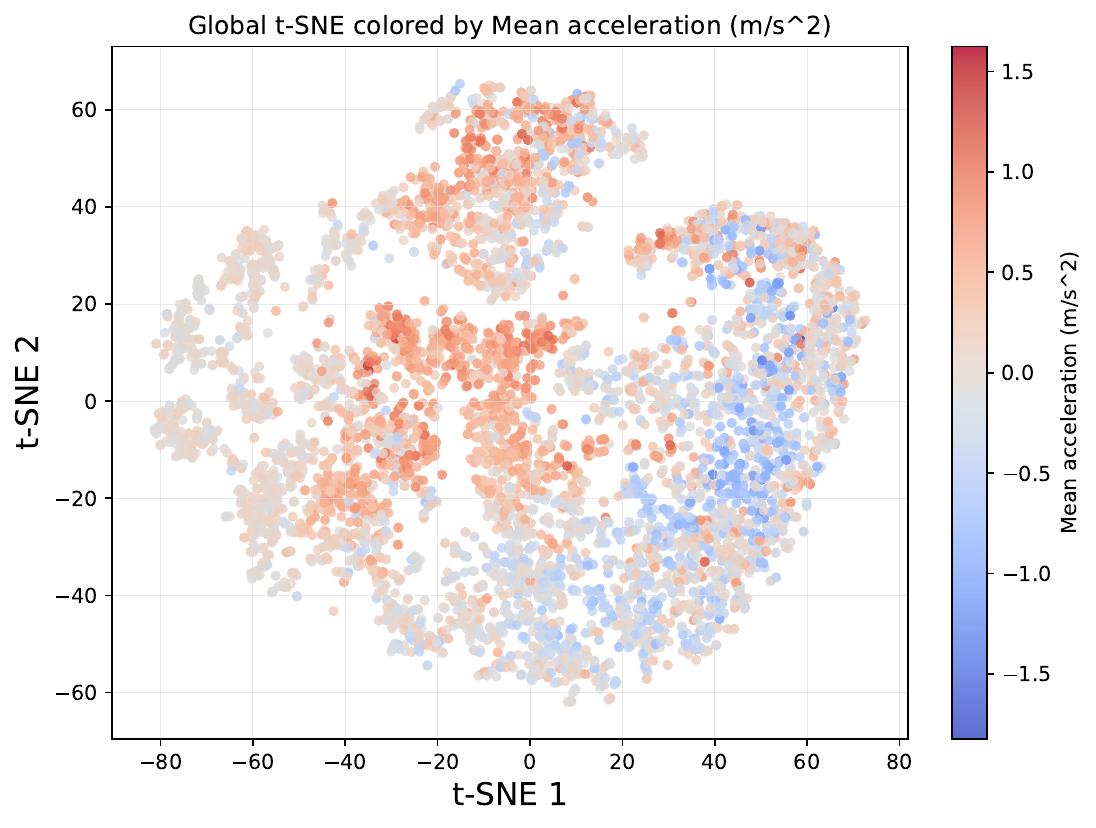}
        \caption{Longitudinal motion}
        \label{fig:latent_longitudinal}
    \end{subfigure}
    \hfill
    \begin{subfigure}[b]{0.32\textwidth}
        \centering
        \includegraphics[width=\textwidth, trim={0mm 0mm 0mm 7mm}, clip]{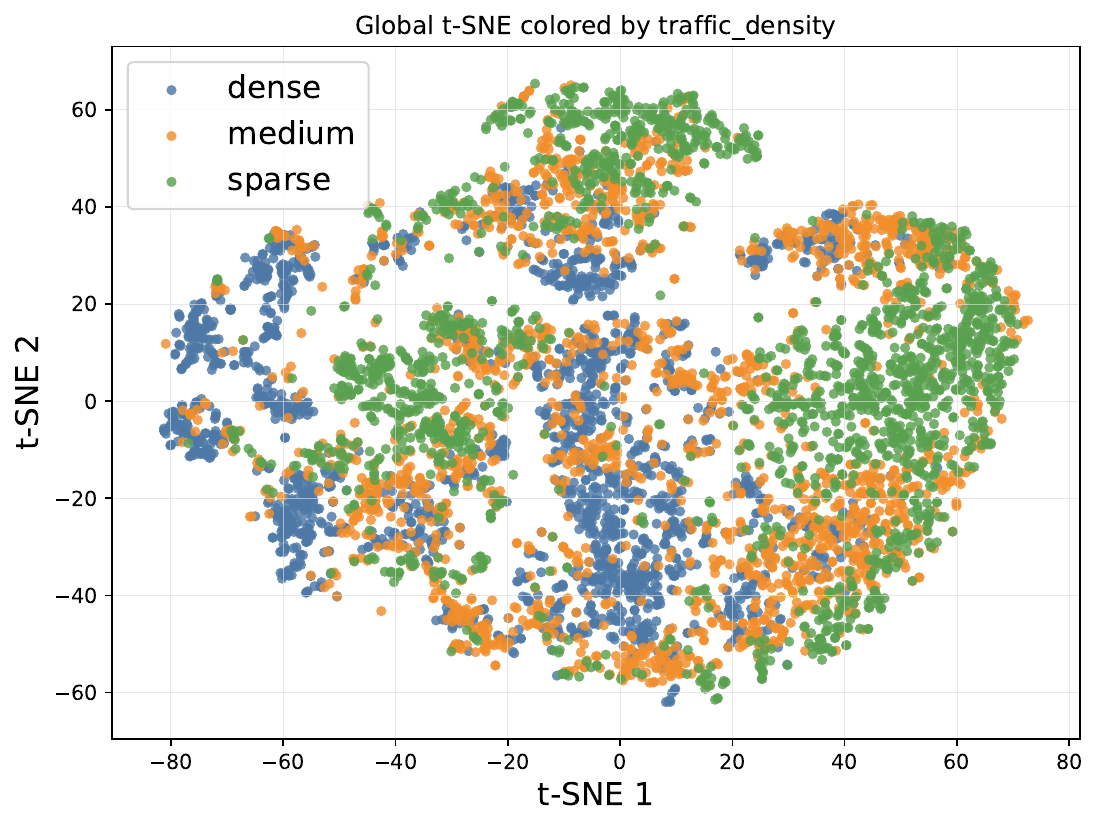}
        \caption{Traffic density}
        \label{fig:latent_density}
    \end{subfigure}
    \caption{
        Visualization of clips in the latent space categorized by rule-based class labels:
        (a)~lateral motion, (b)~longitudinal motion (acceleration), and (c)~traffic density.
        Features form reasonably well-separated clusters across the latent space for each class attribute.
    }
    \label{fig:latent_space_class_labels}
\end{figure}

Furthermore, we analyzed the latent space distributions annotated with rule-based class labels (Figure~\ref{fig:latent_space_class_labels}) to evaluate whether the learned features naturally separate according to high-level semantic categories. The visualization qualitatively demonstrates that the extracted features form discernible clusters, particularly concerning lateral (Figure~\ref{fig:latent_lateral}) and longitudinal motion (Figure~\ref{fig:latent_longitudinal}).

\begin{figure}
    \centering
    \includegraphics[width=1\linewidth, trim={0mm 0mm 0mm 9mm}, clip]{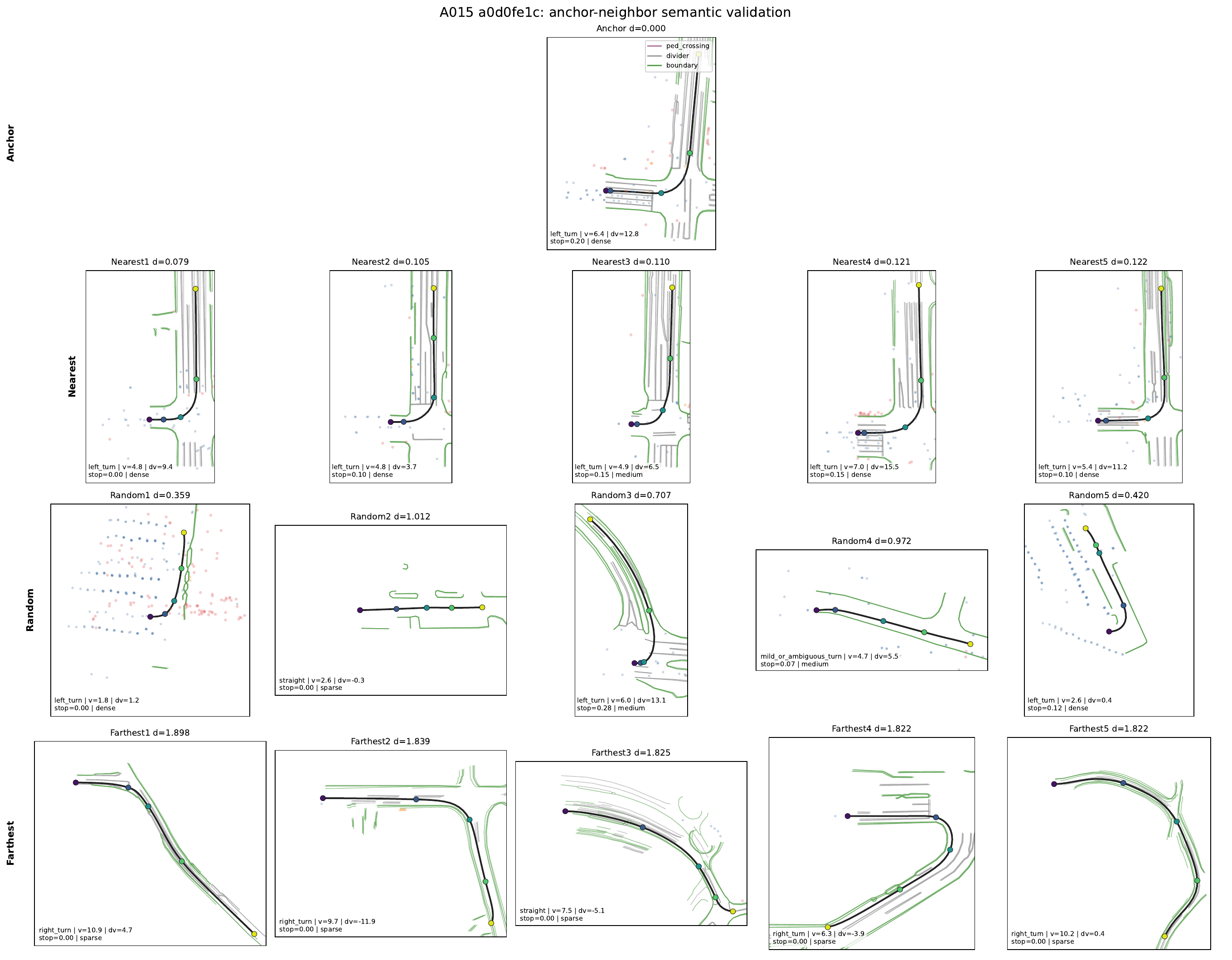}
    \caption{
        Qualitative validation of scene feature distance in the latent space. The top row displays the reference anchor scene ($d=0.000$).
        Subsequent rows show the five nearest ($d \approx 0.079\text{--}0.122$), random intermediate ($d \approx 0.359\text{--}1.012$), and five farthest ($d \approx 1.898\text{--}1.822$) scenes based on distance $d$ in the scene feature space.
        Nearest neighbors preserve key semantic elements—including ego trajectory curvature, road layout geometry, and object density—whereas farthest samples show clear divergence.
    }
    \label{fig:semantic_validation_1}
\end{figure}

To validate that the latent distance metric effectively captures semantic similarity and dissimilarity, we conducted qualitative evaluations comparing an anchor scene against its nearest, random, and farthest neighbors in the latent space (Figure~\ref{fig:semantic_validation_1}). Each scene is represented via a bird's-eye view (BEV) plot detailing the ego trajectory, high-definition (HD) map topology, and surrounding dynamic agents. The results confirm that nearest-neighbor scenes share high semantic similarity in ego trajectory curvature, road layout geometry, and object density, whereas distant samples exhibit fundamentally different driving contexts.
\subsection{HD Map Auto-labeling}
\begin{figure*}[t]
    \centering
    \includegraphics[
        width=\textwidth
    ]{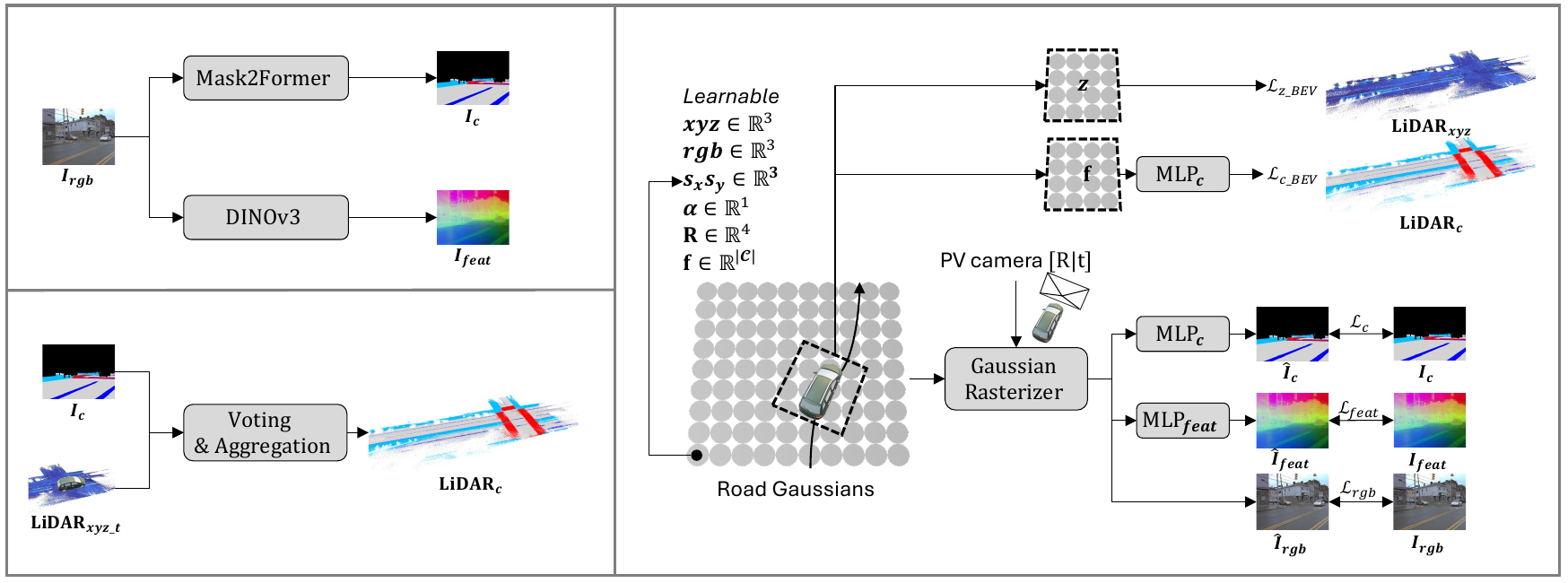}
    \caption{
        Overview of the proposed HD map auto-labeling pipeline.
        A globally aggregated semantic LiDAR point cloud provides
        additional 3D semantic supervision for RoGS-based road
        surface reconstruction. Learnable Gaussian feature embeddings
        are jointly supervised by perspective-view semantic predictions,
        LiDAR-derived semantic labels, and DINOv3 features.
        The reconstructed Gaussian representation is subsequently
        rendered into BEV drivable area pseudo-labels for training
        the End-to-End Autonomous Driving model.
    }
    \label{fig:auto-labeling-pipeline}
\end{figure*}

    To automatically generate HD map annotations on the curated PAI-AV data, we adopt a PMT-based auto-labeling pipeline that employs RoGS~\cite{feng2024rogs} for scene-level road surface reconstruction. We enhance its semantic representation by incorporating LiDAR-derived semantic supervision and DINOv3~\cite{simeoni2025dinov3} feature distillation. The overall auto-labeling pipeline is illustrated in \figureautorefname~\ref{fig:auto-labeling-pipeline}.

\paragraph{Semantic Road Surface Reconstruction.}

RoGS~\cite{feng2024rogs} relies on perspective-view RGB and semantic supervision, which may lead to inconsistent road surface reconstruction due to discrepancies across camera views. To mitigate this issue, we incorporate additional 3D semantic supervision from a globally aggregated static LiDAR point cloud, with semantic consistency improved through multi-view majority voting. Specifically, we project LiDAR points onto multiple camera images and assign semantic labels using Mask2Former~\cite{cheng2022masked} predictions, resolving conflicting labels through majority voting across views. After filtering out dynamic objects and retaining static road-related classes, we transform the labeled points into the global coordinate frame and aggregate them across frames. The resulting semantic point cloud provides additional 3D semantic supervision for road surface reconstruction.

Building upon RoGS~\cite{feng2024rogs}, we expand the Gaussian reconstruction range to 30 m around the historical ego-vehicle trajectory to cover a larger road surface region. To further improve multi-view consistency, we extend the Gaussian representation with learnable feature embeddings jointly supervised by perspective-view semantic predictions, LiDAR-derived semantic labels, and DINOv3~\cite{simeoni2025dinov3} features. Specifically, each Gaussian surfel is parameterized as
\begin{equation}
    \Theta_i =
    \left\{
    (x_i, y_i, z_i),
    (r_i, g_i, b_i),
    (s_{x,i}, s_{y,i}),
    \alpha_i, \mathbf{R}_i, \mathbf{f}_i
    \right\},
    \quad \mathbf{f}_i \in \mathbb{R}^{D},
\end{equation}
where the position, color, scale, opacity, and rotation
follow the original RoGS formulation, and
$\mathbf{f}_i$ denotes a $D$-dimensional learnable
feature embedding associated with the $i$-th Gaussian,
replacing the original semantic vector $\mathrm{Sem}$. 
These embeddings are differentiably rendered into feature maps, which are processed by two separate MLP-based decoders. The semantic decoder predicts semantic labels supervised by both perspective-view semantic predictions and LiDAR-derived semantic labels, while the feature decoder predicts dense DINOv3 features supervised by pretrained DINOv3 representations. This joint supervision encourages the Gaussian embeddings to capture richer and more consistent visual representations across views.

\begin{figure*}[t]
    \centering
    \includegraphics[
        width=\textwidth,
        trim={1mm 2mm 1mm 9mm},
        clip
    ]{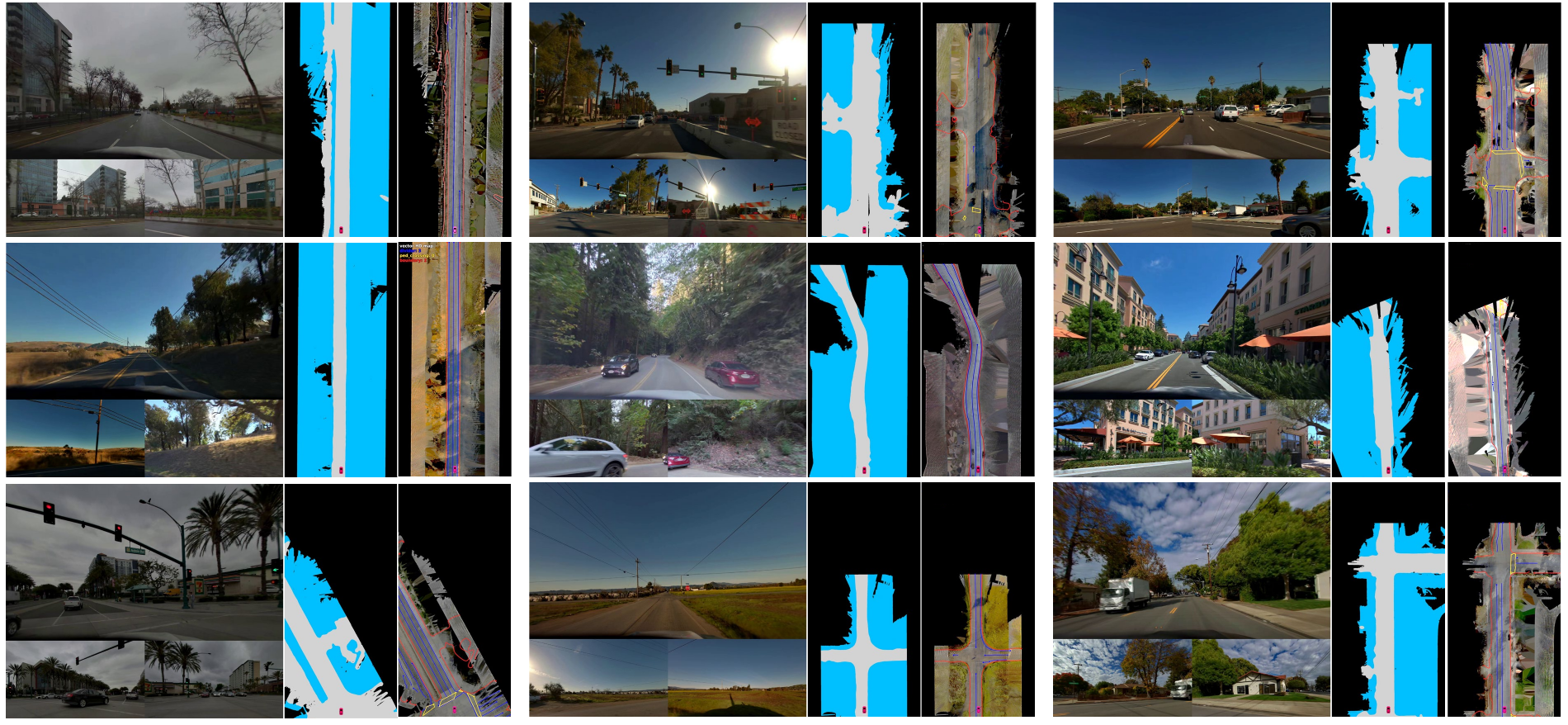}
    \caption{
        Visualization of automatically generated BEV segmentation and vector HD map annotations. Each example shows three perspective-view images on the left (front-wide, cross-left, and cross-right) and the corresponding BEV visualization on the right. The binary segmentation distinguishes drivable and non-drivable areas, while rule-based post-processing generates vectorized map elements comprising dividers (blue), road boundaries (red), and pedestrian crossings (yellow). The BEV region extends 5 m behind and 160 m ahead of the ego vehicle, with a lateral extent of 40 m on each side.
    }
    \label{fig:auto-labeling-results}
\end{figure*}

\paragraph{Semantic HD Map Pseudo-Label Generation.}
Following the PMT auto-labeling pipeline, we generate BEV drivable area pseudo-labels by rendering the reconstructed Gaussian representation within an ego-centric region. Specifically, we define the BEV region to cover 5 m behind and 160 m ahead of the ego vehicle, with a lateral extent of 40 m on each side. Unlike PMT, which considers multiple semantic map categories, we adopt a binary semantic representation that distinguishes only between drivable and non-drivable areas. The resulting BEV segmentation maps serve as pseudo-labels for training the End-to-End Autonomous Driving model. Additionally, we apply rule-based post-processing to the BEV segmentation maps to generate vectorized HD map annotations comprising dividers, road boundaries, and pedestrian crossings. Qualitative examples of the generated BEV segmentation and vector map annotations are presented in \figureautorefname~\ref{fig:auto-labeling-results}.

\subsection{End-to-End Autonomous Driving Model}
    \paragraph{Model Input.}
Along with the camera images, the model takes a projection matrix for each front and front-side camera image, which the BEV encoder uses to map BEV locations to image pixels. It also takes the current frames of the rear-left and rear-right cameras ($512\times256$ each), which are not projected into the BEV. The ego state $\mathbf{s}\in\mathbb{R}^{4}$ consists of the current velocity $(v_x,v_y)$ and acceleration $(a_x,a_y)$; the driving command is not used, as the route replaces it. The model also receives the poses of the two past frames relative to the current rear-axle frame, which are used to align past BEV features with the current frame.

\paragraph{Feature Extractor.}
The DINOv3~\cite{simeoni2025dinov3} ViT-S/16 image backbone encodes each camera image into $32\times16$ patch features at stride 16, which are upsampled and downsampled into a three-level pyramid (strides 8, 16 and 32) with 256 channels per level. In the BEVFormer~\cite{li2024bevformer} encoder, the grid covers 0--56\,m ahead of the vehicle and $\pm28$\,m to each side at 0.5\,m resolution, and each cell is a learnable query that gathers features from this pyramid around its projected locations in each camera that sees it. To save computation, the two past frames pass through only the first encoder layer, without gradients during training, and the encoder outputs a BEV feature map $\mathbf{F}\in\mathbb{R}^{256\times112\times112}$ that is shared by the perception heads and the planner. The drivable-area head applies two ConvNeXt-V2~\cite{woo2023convnext} blocks to $\mathbf{F}$ to predict whether each BEV cell is drivable. The detection head also reads $\mathbf{F}$ and follows the DETR-style BEVFormer head~\cite{carion2020end,li2024bevformer}, with 600 object queries that predict scores for seven classes (vehicle, pedestrian, bicycle, traffic cone, barrier, construction-zone sign and generic object) and a 3D box with position, size, orientation and velocity. The two rear images pass through the same ViT-S one at a time; their $32\times16$ patch maps are placed side by side, projected to 256 channels by a $1\times1$ convolution with GroupNorm and GELU, average-pooled to $32\times8=256$ tokens, and given learned positional embeddings. They use only the current frame, bypass the BEV encoder, and in the final training stage send no gradient to the shared backbone.

\paragraph{Trajectory Vocabulary.}
The vocabulary is built from the 4\,s future trajectory at every frame with a full 4\,s future in the NuRec and PAI-AV clips, sampled as eight poses $(x,y,\theta)$ at 0.5\,s intervals. We first remove trajectories that the MPC controller cannot track. Because turns and lane changes are rare, we add their left--right mirror images, which stay within the controller limits. The trajectories are then interpolated to 40 steps at 0.1\,s and clustered with bisecting k-means, splitting a cluster only if both halves keep at least five trajectories, so every entry averages at least five recorded or mirrored trajectories and satisfies the controller limits.

\begin{figure}[t]
    \centering
    \includegraphics[width=0.99\textwidth]{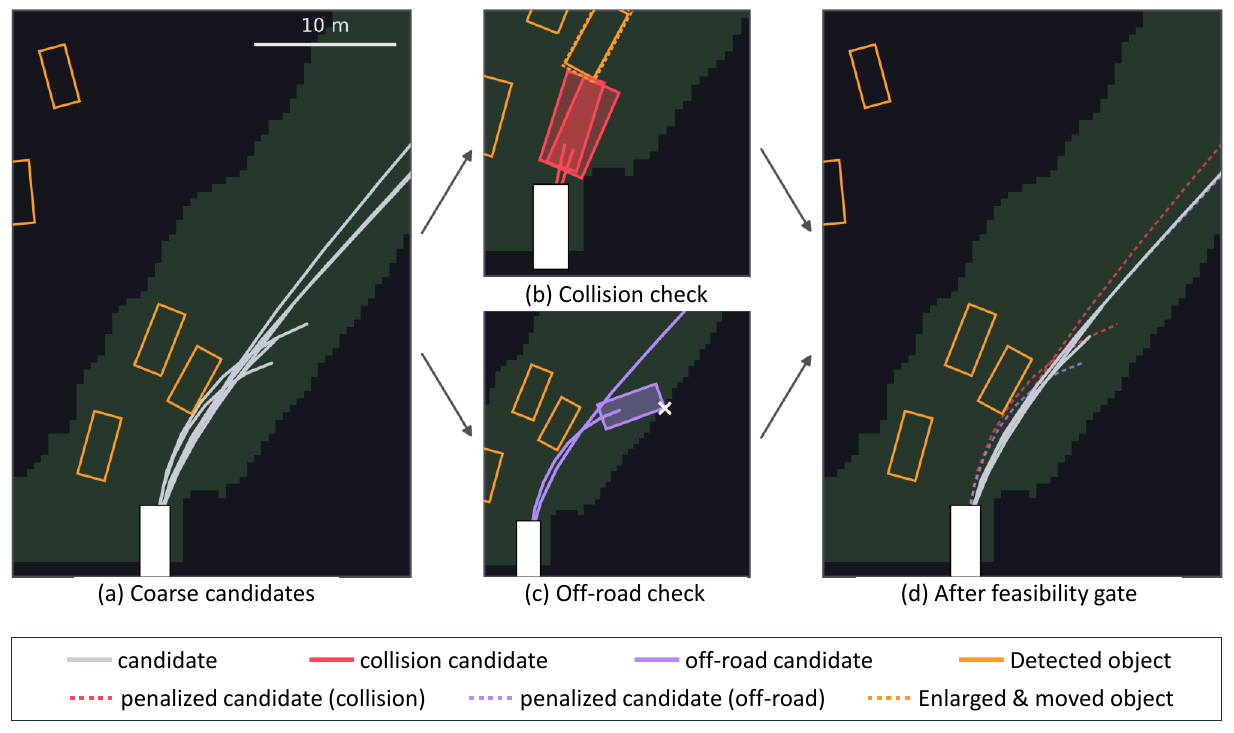}
    \caption{
        Visualization of feasibility gate process: (a) coarse candidates before gating, (b) check the ego footprint against detected objects within the first 1\,s, (c) check the footprint corners against the predicted drivable area, (d) failed candidates penalized and ranked below passing ones.
    }
    \label{fig:feasibility gate}
\end{figure}

\paragraph{Feasibility Gate.}
The gate checks each candidate against the predicted drivable area and detections and returns a binary mask (\figureautorefname~\ref{fig:feasibility gate}). The ego footprint is the clip's rig bounding box (length and width scaled by 0.98), placed ahead of the rear-axle pose by the rear-axle-to-center offset and enlarged by 1.1. A candidate fails if, at any of its 40 poses, a footprint corner inside the BEV range falls on a cell with drivable probability below 0.6, or if, within its first 10 poses (1\,s), the footprint overlaps a detection with confidence above 0.4 under the separating axis theorem; detections are enlarged by 2.0 for pedestrians and bicycles and 1.1 otherwise, and moved at their predicted velocity with fixed heading. Both thresholds were chosen to maximize segmentation IoU and detection F1 on validation. If fewer than 256 candidates pass, the highest-scoring failed ones fill the remaining slots.

\paragraph{Planner.}
In the coarse stage, each vocabulary trajectory (40 poses $(x,y,\theta)$) is flattened, embedded by an MLP, and passed through one self-attention layer over all candidates to form the decoder queries. The keys and values are the pooled BEV tokens with learned positional embeddings and the 256 rear tokens, concatenated with 20 route tokens (one per waypoint, with empty slots masked). The ego state and a route summary are added to each candidate's decoder output, from which MLP heads predict NC, DAC, EP, the overall score and imitation. The coarse score is the product of the predicted NC, DAC and EP plus the imitation score normalized to $[0,1]$, all with weight 1.0; the overall-score head is trained but not used for ranking. Candidates failing the feasibility gate are demoted by the score range plus one, and the top 256 are kept. In the fine stage, their decoder outputs serve as queries of a three-layer refinement decoder with its own heads, attending to the pooled BEV tokens, the rear tokens and its own route tokens. All three layers are supervised, and the final trajectory is the candidate with the highest last-layer score, computed as in the coarse stage but with an imitation weight of 0.02.

\paragraph{Controller.}
\label{app_controller}
Planner labels are generated with a batched re-implementation of the AlpaSim linear MPC controller, so they reflect how the simulator would drive each candidate. The vehicle is a dynamic bicycle model (kinematic below 5\,m/s) with a 2.85\,m wheelbase, whose steering and acceleration follow their commands through first-order lags of 0.1\,s. Every 0.1\,s, the controller linearizes the model about the current state and solves a quadratic program over a 2\,s horizon, penalizing longitudinal, lateral and heading errors (weights 2.0, 1.0 and 1.0), the acceleration state (0.1), and the steering and acceleration commands (5.0 and 1.0), with tracking applied only from 1\,s ahead. The states are bounded by a heading change of at most $90^\circ$, $0 \le v_x \le 35$\,m/s, a steering angle of at most $45^\circ$, and $-8 \le a \le 6$\,m/s$^2$. Each of the 8,192 vocabulary trajectories is simulated for 4\,s from the recorded ego state, re-solving the MPC at every step.

\subsection{Data-driven Imitation Learning}
    \label{app_il}

\paragraph{Data Filtering.}
We remove clips with unreliable labels or duplicated drives. For NuRec, these are clips with reverse driving that the MPC cannot follow, missing or inconsistent map or route geometry, or object boxes unsupported by any camera frame. For PAI-AV, these are clips in which the ego vehicle leaves the annotated drivable area, along with clips that duplicate a NuRec drive.

\paragraph{Perception Loss.}
BEV segmentation uses class-weighted cross-entropy with inverse-frequency class weights and a loss weight of 4.5. Detection follows the BEVFormer head~\cite{li2024bevformer}: a sigmoid focal loss~\cite{lin2017focal} ($\gamma=2$, $\alpha=0.25$) for classification with weight 2.0 and an L1 loss on the 3D box parameters with weight 0.25. Queries are assigned by Hungarian matching~\cite{carion2020end} on the same weighted sum of the focal classification cost and the L1 box cost, and every decoder layer is supervised.

\paragraph{Planning Loss.}
The imitation and score losses follow DriveSuprim~\cite{yao2026drivesuprim}. The imitation target is a softmax over negative squared distances to the recorded trajectory at eight timestamps (0.5\,s apart), with temperature $\sigma=0.5$. The loss weights are 1.0 for imitation and 3.0 (NC), 3.0 (DAC), 2.0 (EP), and 1.0 (their product) for the score heads. Scenes with identical labels for all candidates, mostly beyond the MPC speed limit, are excluded from the score losses.

\paragraph{Training Stage.}
Stage 1 initializes the image backbone from DINOv3~\cite{simeoni2025dinov3} and disables the planning losses; each later stage initializes from the previous one with a fresh optimizer and schedule. We keep DriveSuprim's~\cite{yao2026drivesuprim} EMA teacher, which is used at inference, and student-only photometric augmentation, with momentum rising from 0.992 to 0.998 over the first three epochs, but do not use its rotation-augmented views. The feasibility gate is also applied during Stages 2 and 3 using predicted detections and drivable area rather than ground truth, so the fine stage is trained on the candidate distribution it receives at inference. Training uses bf16 mixed precision and gradient clipping at a global norm of 35.
\subsection{Simulator-based Reinforcement Learning}
    \label{subsec:implementation_details}
\label{app_RL}

\paragraph{Policy Architecture and Optimization.}
We train our policy using Proximal Policy Optimization (PPO) \cite{schulman2017proximal} and Generalized Advantage Estimation (GAE) \cite{schulman2018highdimensionalcontinuouscontrolusing}. The actor is parameterized via Low-Rank Adaptation (LoRA), while the critic is trained jointly. Both networks are optimized using AdamW ($\beta_1 = 0.9$, $\beta_2 = 0.999$, $\epsilon = 10^{-5}$) with a constant learning rate of $10^{-4}$. Weight decay is set to $0.01$ for LoRA parameters.

\paragraph{Experience Rollout and Distributed Training.}
Each update collects a buffer of about 2,048 decision steps from 48 concurrent simulation sessions. Each update runs up to 3 PPO epochs over 4 minibatches of 512 steps, i.e. up to 12 optimizer steps. Training uses 8× NVIDIA H100 (80 GB) GPUs with data parallelism. Each GPU hosts one learner rank, which also serves policy inference for the sessions it owns, together with seven neural-rendering instances and a physics service. Each rank processes about 64 steps per minibatch with a micro-batch size of 8 via gradient accumulation.

\paragraph{RL Objectives and Regularization.}
To align with the $10$\,Hz decision frequency, we set discount $\gamma = 0.951$ ($0.99^5$) and GAE $\lambda = 0.774$ ($0.95^5$). Policy clip $\epsilon$, value clip threshold, and value loss coefficient are set to $0.1$, $0.1$, and $0.5$, respectively. To prevent deviation from the SFT baseline, an exact KL penalty ($\beta_{\text{KL}} = 0.02$) is computed over $256$ candidates, and training early-stops if approximate KL exceeds $1.5 \times 0.03$. Gradients are clipped independently at norm $0.5$. Advantage normalization is executed per minibatch across all GPUs. Policy entropy is regulated via a sampling temperature targeting $2.5$\,nats without explicit entropy loss.

\begin{table}[t]
\centering
\caption{Reinforcement learning hyperparameters and system configuration.}
\label{tab:hyperparameters}
\small
\begin{tabular}{ll}
\toprule
\textbf{Parameter / Setting} & \textbf{Value} \\
\midrule
Algorithm & PPO\\
Optimizer & AdamW ($\beta_1=0.9, \beta_2=0.999, \epsilon=10^{-5}$) \\
Learning Rate & $10^{-4}$\\
Weight Decay & $0.01$ \\
PPO Epochs & 12 (1st stage), 5 (2nd stage)\\
Batch Size & $2048$ \\
Discount ($\gamma$) / GAE ($\lambda$) & $0.951$ / $0.774$ \\
Clip ($\epsilon$) / Value Clip & $0.1$ / $0.1$ \\
Value Loss Coef. & $0.5$ \\
KL Penalty ($\beta_{\text{KL}}$) & $0.02$ \\
Early Stopping & Approx. KL $> 0.045$ \\
Gradient Clipping & $0.5$ ($L_2$ norm) \\
Advantage Normalization & Per minibatch \\
Hardware & $8 \times$ NVIDIA H100 80\,GB \\
Training Overhead & $12$ + $5$ epochs ($460$ updates, $\sim 34.5$\,h) \\
\bottomrule
\end{tabular}
\end{table}

\paragraph{Implementation Details.}
The complete training run consists of two stages. The first stage spans $12$ epochs over $2{,}202$ scenes, comprising $26{,}424$ episodes and $391$ PPO updates, and completes in approximately $29.5$ hours. The second stage continues from this checkpoint for $5$ additional epochs on the $894$ scenes with mixed success and failure outcomes, adding $4{,}470$ episodes and $69$ PPO updates in approximately $5$ hours. In total, training comprises $30{,}894$ episodes and $460$ PPO updates over approximately $34.5$ hours of wall-clock time on the $8 \times$ H100 setup. Detailed hyperparameter configurations are summarized in Table~\ref{tab:hyperparameters}.
\end{document}